%% file: main_arxiv.tex
\documentclass[10pt,twocolumn,letterpaper]{article}

\usepackage[pagenumbers]{cvpr} % To force page numbers, e.g. for an arXiv version
\usepackage[accsupp]{axessibility}

\usepackage{graphicx}
\usepackage{amsmath}
\usepackage{amssymb}
\usepackage{booktabs}
\usepackage{makecell}
\usepackage{multicol}

\usepackage{multirow}
\usepackage{comment}
\usepackage{color, colortbl}
\usepackage{times}
\usepackage{epsfig}
\usepackage{graphicx}
\usepackage{amsmath}
\usepackage{amssymb}
\usepackage[percent]{overpic}
\usepackage{times}
\usepackage{epsfig}
\usepackage{graphicx}
\usepackage{float}
\usepackage{wrapfig}
\usepackage{amsmath,amssymb,amsthm}
\usepackage{verbatim}
\usepackage{multirow}
\usepackage{balance}
\usepackage{url}
\usepackage{booktabs}
\usepackage{etoolbox}
\usepackage{pifont,hologo}
\usepackage{dblfloatfix}
\usepackage{soul}
\usepackage{pifont}
\usepackage{wasysym}

\makeatletter
\@namedef{ver@everyshi.sty}{}
\makeatother

\usepackage{tikz}

\input{custom}

\usepackage[pagebackref,breaklinks,colorlinks]{hyperref}

\usepackage[capitalize]{cleveref}
\crefname{section}{Sec.}{Secs.}
\Crefname{section}{Section}{Sections}
\Crefname{table}{Table}{Tables}
\crefname{table}{Tab.}{Tabs.}

\newcommand\blfootnote[1]{%
  \begingroup
  \renewcommand\thefootnote{}\footnote{#1}%
  \addtocounter{footnote}{-1}%
  \endgroup
}

\def\cvprPaperID{7771} % *** Enter the CVPR Paper ID here
\def\confName{CVPR}
\def\confYear{2023}

\begin{document}

%%%%%%%%% TITLE - PLEASE UPDATE
\title{ReLight My NeRF: A Dataset for \\Novel View Synthesis and Relighting of Real World Objects}

\author{
Marco Toschi\thanks{Joint first authorship. Work done while at Eyecan.ai}, Riccardo De Matteo$^{*,\diamond}$, Riccardo Spezialetti$^*$, Daniele De Gregorio\\
Eyecan.ai\\
\texttt{\small\{marco.toschi, riccardo.spezialetti, daniele.degregorio\}@eyecan.ai} \\
%\AND
Luigi Di Stefano, Samuele Salti\\
University of Bologna
}

\twocolumn[{
\renewcommand\twocolumn[1][]{#1}
\maketitle
\begin{center}
    \vspace{-12.00mm} 
    \centering
    \captionsetup{type=figure}
    \includegraphics[width=1\textwidth]{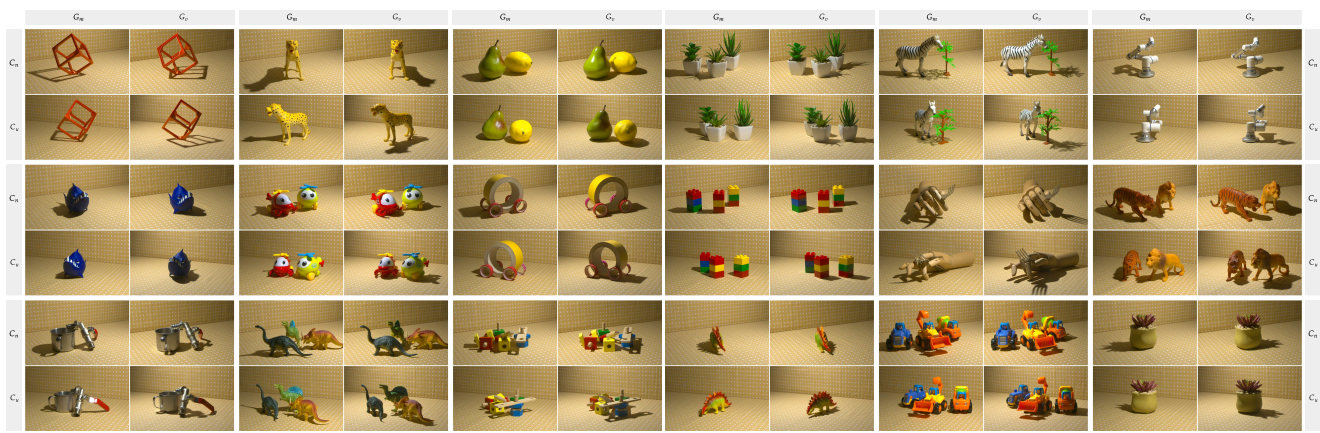}
    \captionof{figure}{\textbf{A random subset of scenes from our dataset}. Our dataset enables the study of joint novel view synthesis and relighting from real images, as it provides scenes framed from the same viewpoint under varying light positions (shown along rows) as well as captured from different viewpoints under the same light position (shown along columns).} 
    \label{fig:teaser}    
\end{center}
}]

%%%%%%%%% ABSTRACT
\begin{abstract}
In this paper, we focus on the problem of rendering novel views from a Neural Radiance Field (NeRF) under unobserved light conditions. To this end, we introduce a novel dataset, dubbed \datasetname{} (\textbf{Re}lighting \textbf{Ne}RF), framing real world objects under \textit{one-light-at-time} (OLAT) conditions, annotated with accurate ground-truth camera and light poses. Our acquisition pipeline leverages two robotic arms holding, respectively, a camera and an omni-directional point-wise light source. We release a total of 20 scenes depicting a variety of objects with complex geometry and challenging materials. Each scene includes $2000$ images, acquired from 50 different points of views under 40 different OLAT conditions. By leveraging the dataset, we perform an ablation study on the relighting capability of variants of the vanilla NeRF architecture and identify a lightweight architecture that can render novel views of an object under novel light conditions, which we use to establish a non-trivial baseline for the dataset. Dataset and benchmark are available at \url{https://eyecan-ai.github.io/rene}.
\end{abstract}

\input{section/introduction.tex}

\input{section/related.tex}
\input{section/dataset}

\input{section/benchmark}
\input{section/conclusion}
%%%%%%%%% REFERENCES
{\small
\bibliographystyle{ieee_fullname}
\bibliography{egbib}
}

\end{document}

%% file: custom.tex
\newcommand{\datasetname}{ReNe}

\def\eg{\emph{e.g}\onedot} 
\def\ie{\emph{i.e}\onedot}

\newcommand{\image}{I}
\newcommand{\imageset}{\mathcal{I}}

\newcommand{\sceneset}{\mathcal{S}}
\newcommand{\camera}{C}
\newcommand{\camerasset}{\mathcal{C}}
\newcommand{\light}{G}
\newcommand{\lightsset}{\mathcal{G}}
\newcommand{\rotation}{\mathbf{R}}
\newcommand{\translation}{\textbf{t}}
\newcommand{\cmark}{\ding{51}}
\newcommand{\xmark}{\ding{55}}

\newcommand{\camerabot}{\emph{CameraBot}}
\newcommand{\lightbot}{\emph{LightBot}}
\newcommand{\rgbnet}{\Psi_{rgb}}
\newcommand{\geonet}{\Psi_{geo}}
\newcommand{\shanet}{\Psi_{vis}}
\newcommand{\x}{\mathbf{x}}
\newcommand{\dir}{\mathbf{d}}
\newcommand{\ldir}{\mathbf{l}}
\newcommand{\cl}{c}
\newcommand{\geoemb}{\mathbf{e}}
\newcommand{\sh}{\zeta}
\newcommand{\hash}{h}

%% file: section/introduction.tex
\section{Introduction}
\label{sec:intro}
\blfootnote{$^*$ Joint first authorship. $^\diamond$ Work done while at Eyecan.ai}
\textit{Inverse rendering} \cite{sato1997object,marschner1998inverse,yu1999inverse,ramamoorthi2001signal} addresses the problem of estimating the physical attributes of an object, such as its geometry, material properties and lighting conditions, from a set of images or even just a single one. This task is a longstanding problem for the vision and graphics communities, since it unlocks the creation of novel renderings of an object from arbitrary viewpoints and under unobserved lighting conditions. An effective and robust solution to this problem would have significant value for a wide range of applications in gaming, robotics and augmented reality. 
% add citations

Recently, Neural Radiance Fields (NeRF)\cite{mildenhall2020nerf} has contributed tremendously to the novel view synthesis sub-task of inverse rendering pipelines. By mapping an input 5D vector (3D position and 2D viewing direction) to a 4D continuous field of volume density and color by means of a neural network, NeRF learns the geometry and appearance of a single scene from a set of posed images.
The appealing results in novel view synthesis have attracted a lot of attention from the research community and triggered  many follow-up  works aimed at overcoming the main limitations of NeRF, \eg{} reduce inference runtime \cite{liu2020neural,lindell2021autoint,rebain2021derf,reiser2021kilonerf,yu2021plenoctrees,lombardi2021mixture}, enable modeling of deformable objects \cite{park2021nerfies,pumarola2021d,gafni2021dynamic,liu2021neural,tretschk2021non,peng2021animatable}, and generalization to novel scenes \cite{schwarz2020graf,trevithick2021grf,yu2021pixelnerf,tancik2021learned,chan2021pi,rematas2021sharf,niemeyer2021campari,kosiorek2021nerf,jang2021codenerf}. However, less attention has been paid to the relighting ability of NeRFs. Although NeRF and its variants represent nowadays the most compelling strategy for view synthesis, the learned scene representation entangles material and lighting and, thus, cannot be directly used to generate views under novel, unseen lighting conditions.  A few existing works \cite{bi2020deep,srinivasan2021nerv,zhang2021nerfactor,boss2021nerd} try to overcome this structural NeRF limitation by learning to model the scene appearance as a function of \textit{reflectance}, which accounts for both scene geometry and lighting modeling. However, these methods incur a great computation cost, mainly due to the need to explicitly model light visibility and/or the components of a microfacet Bidrectional Reflectance Distribution Function (BRDF) \cite{walter2007microfacet} like diffuse albedo, specular roughness, and point normals: for instance NeRV \cite{srinivasan2021nerv} uses 128 TPU cores for 1 day while \cite{bi2020deep} is trained on 4 GPUs for 2 days.

\begin{table}[t]
    \resizebox{\linewidth}{!}{
      \centering
      \setlength{\tabcolsep}{1.5mm}
      {\small
       \begin{tabular}{l|ccccc}
       \toprule
       Dataset & \thead{Multiple\\categories}  & Real-World & \thead{Background\\Shadows} & Public & \thead{Light \\ Supervision} \\
       \midrule       
       Gross \etal{} \cite{gross2010multi}          & \xmark & \cmark & \xmark & \cmark & \cmark\\
       Sun \etal{} \cite{sun2019single}             & \xmark & \cmark & \xmark & \cmark & \cmark\\       
       Wang \etal{} \cite{wang2020single}           & \xmark & \cmark & \xmark & \cmark & \cmark\\
       Zhang \etal{} \cite{zhang2021neural}         & \xmark & \cmark & \xmark & \cmark & \cmark\\
       Srinivasan \etal{} \cite{srinivasan2021nerv} & \cmark & \xmark & \xmark & \cmark & \cmark\\
       Zhang \etal{} \cite{zhang2021nerfactor}      & \cmark & \cmark & \cmark & \cmark & \xmark\\
       Zhang \etal{} \cite{zhang2021nerfactor}      & \cmark & \xmark & \xmark & \cmark & \cmark\\
       Bi \etal{} \cite{bi2020neural}               & \cmark & \cmark & \xmark & \xmark & \cmark\\
     \bottomrule
     \datasetname{} & \cmark & \cmark & \cmark & \cmark & \cmark \\ 
     \midrule       
      \end{tabular}
      }
   }
  %\vspace{3mm}
  \caption{\label{tab:dataset} Overview of relighting datasets. Our dataset is the first one featuring a variety of objects and materials captured with real-world sensors that provides ground-truth light positions and also presents challenging cast shadows.}
\end{table}

As highlighted in \cref{tab:dataset}, one of the main challenges to foster research in this direction is the absence of real-world datasets featuring generic and varied objects with ground-truth light direction, both at training and test time. The latter is key to create a realistic quantitative benchmark for relighting methods, whose availability is one of the main driving forces behind fast-paced development of a machine learning topic. Indeed, the above mentioned NeRF-like methods for relighting mainly consider a handful of synthetic images to provide quantitative results (3 and 4 scenes in NeRV \cite{srinivasan2021nerv} and NeRFactor \cite{zhang2021nerfactor}, respectively) while no quantitative results are provided in \cite{bi2020deep}. The only dataset with scenes acquired by a real sensor is proposed in \cite{bi2020neural}, which, however, assumes images captured under collocated view and lighting setup, \ie{} a smartphone with flash light on, which limits the amount of cast shadows and simplifies the task. Moreover, the dataset is not publicly available. Some available real-world datasets with ground-truth light positions feature human faces or portraits \cite{gross2010multi,sun2019single,wang2020single,zhang2021neural}. In these datasets, the subject is usually seated in the center of a light-stage with cameras arranged over a dome array positioned in front of the subject. Although these dataset provide real scenes with ground-truth annotations for light positions, %the variability of the scenes is limited and they lack complex geometries and reflective surfaces. Moreover, 
the background is masked-out and shadows cast on the background, which are hard to model because they require a precise knowledge about the geometry of the overall scene, are ignored.

Therefore, in this paper we try to answer the research questions: can we design a data acquisition methodology suitable to collect a set of images of an object under \textit{one-light-at-time} (OLAT) \cite{zhang2021neural} illumination with high-quality camera and light pose annotations which requires minimal human supervision? We then leverage it to investigate a second question: can we design a novel Neural Radiance Field architecture to learn to perform relighting with reasonable computational requirements? 
To answer the first question, we design a capture system relying on two robotic arms. While one arm holds the camera and shots pictures from viewpoints uniformly distributed on a spherical wedge, the other moves the light source across points uniformly distributed on a dome. %Thanks to this setup, we do not need to rely on classical structure-from-motion algorithm to estimate the camera or light pose \cite{schonberger2016structure}. 
As a result, we collect the ReNe dataset, made of $20$ scenes framing daily objects with challenging geometry, varied materials and visible cast shadows, composed of $50$ camera view-points under $40$ OLAT light conditions, i.e. $2000$ frames per scene. Examples of images from the dataset are shown in Fig. \hyperref[fig:teaser]{1}. With a subset of images from each scene, we create a novel hold-out dataset for joint relighting and novel view synthesis evaluation that will be used as an online benchmark to foster research on this important topic. 
As regards the second question, thanks to the new dataset we conduct a study on the relighting capability of NeRF. In particular, we investigate on how the standard NeRF architecture can be modified to take into account the position of the light when generating the appearance of a scene. Our study shows that by estimating color with two separate sub-networks, one in charge of soft-shadow prediction and one responsible for neurally approximating  the BRDF, we can perform an effective relighting, \eg{} cast complex shadows. We provide results of our novel architecture as a reference baseline for the new benchmark.

In summary, our contributions include:
\begin{itemize}
    \item a novel dataset made out of sets of OLAT images of real-world objects, with accurate camera and light pose annotations;
    \item a study comparing different approaches to enable NeRF to perform relighting alongside  novel view synthesis;
    \item a new architecture, where the stage responsible for radiance estimation is split into two separate networks, that can render novel views under novel unobserved lighting conditions;
    \item a public benchmark for novel view synthesis and relighting of real world objects, that will be maintained on an online evaluation server.
\end{itemize}

%% file: section/related.tex
\section{Related Work}
\label{sec:related}
In this paper, we focus on relighting static objects by Neural Radiance Fields.
We briefly discuss the related works and their datasets below.

\noindent\textbf{Image Relighting without NeRF.}
There exist several works in the field of image based relighting \cite{debevec2000acquiring,matusik2004progressively,peers2009compressive,ren2015image}. The proposed methods differ for the adopted technique to model the light transport function in form of discrete light transport matrix. With recent advances in deep learning, new techniques have been introduced to address relighting \cite{sun2019single}, with many of them designed to relight human portraits \cite{xu2018deep,zhou2019deep,nestmeyer2020learning,sun2020light,sun2021nelf}. Another bunch of works pursues  joint relighting and novel view synthesis \cite{meka2020deep,gao2020deferred,zhang2021neural}.

\noindent\textbf{Relighting Datasets.}
Most of relighting datasets contain captures of human bodies or face portraits, they leverage complex capture setups such as calibrated multi-view light-stages with cameras and LED lights, where the cameras are synchronized with the lights in order to flash one LED per capture \cite{nestmeyer2020learning,sun2020light,zhou2019deep,sun2021nelf}. Alternative works on generic objects do exist \cite{xu2018deep,ren2015image}, but they consider a mix of real and synthetic data, with the latter dominating the former in terms of number of images. The dataset most similar to ours is the one adopted by \cite{bi2020neural}, which involves a robotic arm setup holding a Samsung Galaxy Note 8. However, this acquisition framework assumes a collocated view and lighting setup, \eg built-in flash of a smart-phone camera lens, that inherently limits the shadows cast. Furthermore, the dataset has not been released.

\noindent\textbf{Neural Radiance Fields.} Novel view synthesis has been a longstanding problem within the computer vision and computer graphics fields \cite{levoy1996light,gortler1996lumigraph,davis2012unstructured}. The advent of deep learning gave rise to explicit methods that train CNNs for this very purpose\cite{zhou2018stereo,mildenhall2019local, srinivasan2019pushing,li2020crowdsampling,tucker2020single, lombardi2019neural,sitzmann2019deepvoxels}. In the past few years, NeRF has advanced greatly in becoming the main scene representation for view synthesis\cite{mildenhall2020nerf}. NeRF learns a continuous volumetric function parameterized by means of a fully connected neural network optimized over a set of observed images with known camera poses using gradient descent. Due to its effectiveness, NeRF has inspired many subsequent works that extend its continuous neural volumetric representation for generative modeling \cite{chan2021pi,schwarz2020graf,kosiorek2021nerf}, dynamic scenes \cite{li2021neural,ost2021neural,martin2021nerf, pumarola2021d, li2021neural, xian2021space, gao2021dynamic}, non-rigidly deforming objects \cite{gafni2021dynamic,park2021nerfies, tretschk2021non, gafni2021dynamic, noguchi2021neural, park2021hypernerf}, multi-resolution images \cite{barron2021mip,takikawa2021neural}, phototourism images with changing illumination settings \cite{tancik2021learned,martin2021nerf} and relighting \cite{bi2020neural,boss2021nerd,srinivasan2021nerv,martin2021nerf}.

\noindent\textbf{Neural Radiance Fields for Relighting.} One of main limitations of NeRF is that is not suitable for relighting. Indeed, NeRF treats the particles within scene representation as elements that \textit{emit} light instead of being modelled as particles that \textit{reflect} the incoming light sent out from external light sources. To straddle this divide, Neural Reflectance Field \cite{bi2020neural} models both scene geometry and \textit{reflectance} regressing volume density, normal and material properties, \eg  Bidirectional Reflectance Distribution Function (BRDF) \cite{walter2007microfacet}, for any 3D location within the volume of a scene. The shading at each point is then estimated using the light, view direction, the normal at that point and the BRDF. For instance, \cite{lyu2022neural} leverages a differentiable path tracer to optimize a spatially-varying BSDF tied to an the environment map, that can be used to render novel views under a novel OLAT condition. Unfortunately, this approach requires marching rays from all points sampled along the camera ray to any source of lighting in the scene, which restricts it use only in a collocated camera-light setup. NeRV ameliorates the computational cost of \cite{bi2020neural} by replacing the visibility estimation between scene points and light sources with a neural approximation of the true visibility field, which acts as a lookup table during rendering. Thanks to this insight, NeRV simulates direct illumination from environment lighting as well as one-bounce indirect illumination. 
NeRD \cite{boss2021nerd} does not model visibility or shadows, but uses an analytic BRDF model \cite{cook1981reflectance} to learn a volumetric representation which stores SVBRDF \cite{kajiya1986rendering} parameters at each 3D point instead of a radiance field. A relightable textured mesh is then extracted from that volume to allow fast rendering and relighting. Finally, NeRFactor \cite{zhang2021nerfactor} 
starts from two pre-trained networks: a NeRF of the scene and a BRDF network trained on the MERL dataset \cite{matusik2003data}. This knowledge is then distilled into four MLP-based networks to predict for each surface location its normal vector, light visibility, albedo and a BRDF latent code. The MLP outputs are injected into classical volumetric rendering and the networks parameters are optimized minimizing the re-rendering loss. 
\\
We take inspiration from this line of works to design the study reported in \cref{sec:benchmark}, where we split the  MLP-stack of NeRF in two separate networks: one responsible for the visibility and one that resembles the BRDF.

%% file: section/dataset.tex
\section{Dataset Acquisition Framework}
\label{sec:dataset_acquisition}

\begin{figure*}
    \centering
    \begin{subfigure}[b]{0.56\textwidth}
        \includegraphics[width=\textwidth]{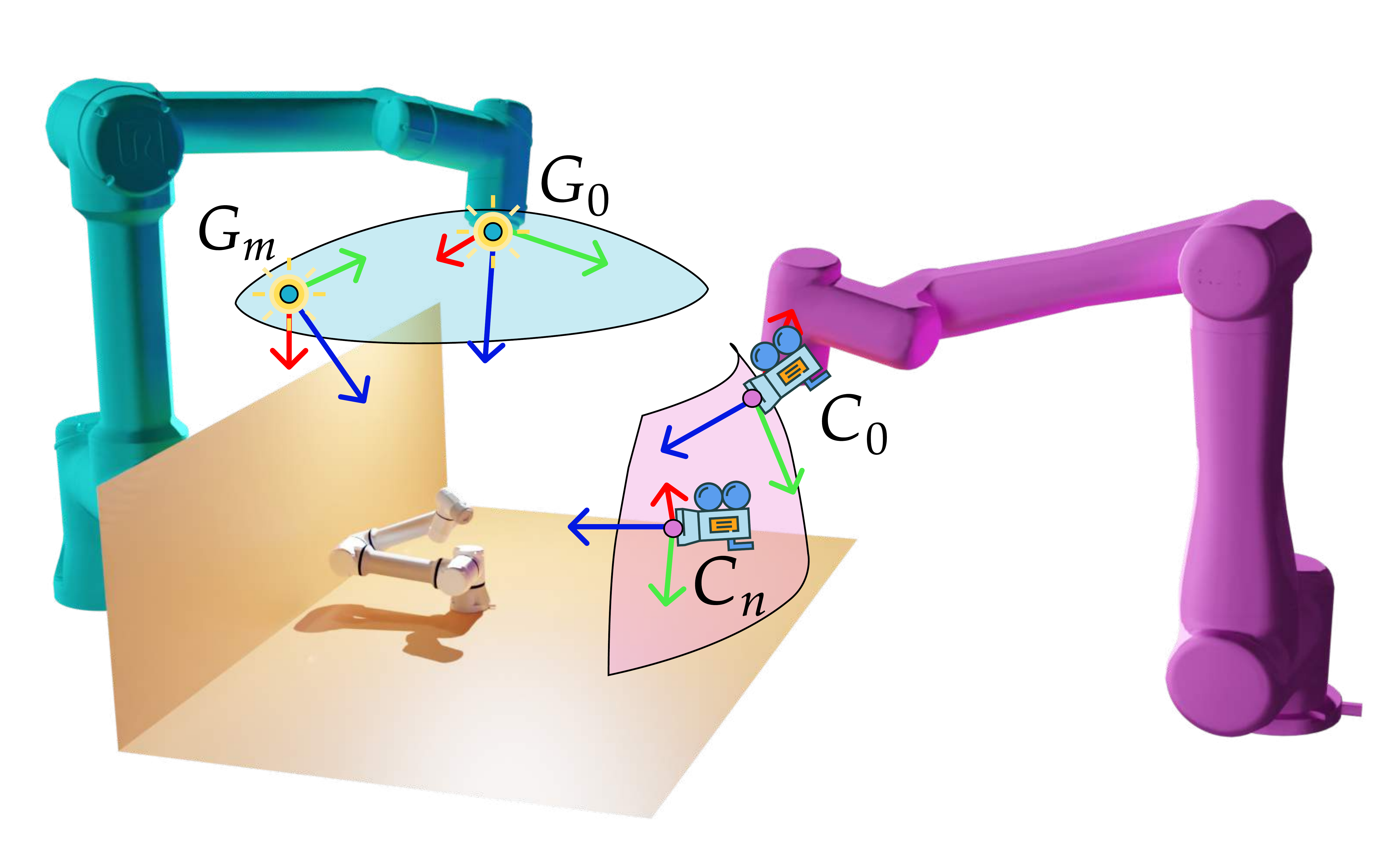}
        \caption{}
        % \caption{The \lightbot{} and \camerabot{} moving light and camera respectively; the trajectories of the two robots are two non-intersecting sections of a hemisphere around the object of interest.}
        \label{fig:hardware_setup}
    \end{subfigure}
    \hfill
    \begin{subfigure}[b]{0.41\textwidth}
        \centering
        \includegraphics[width=\textwidth]{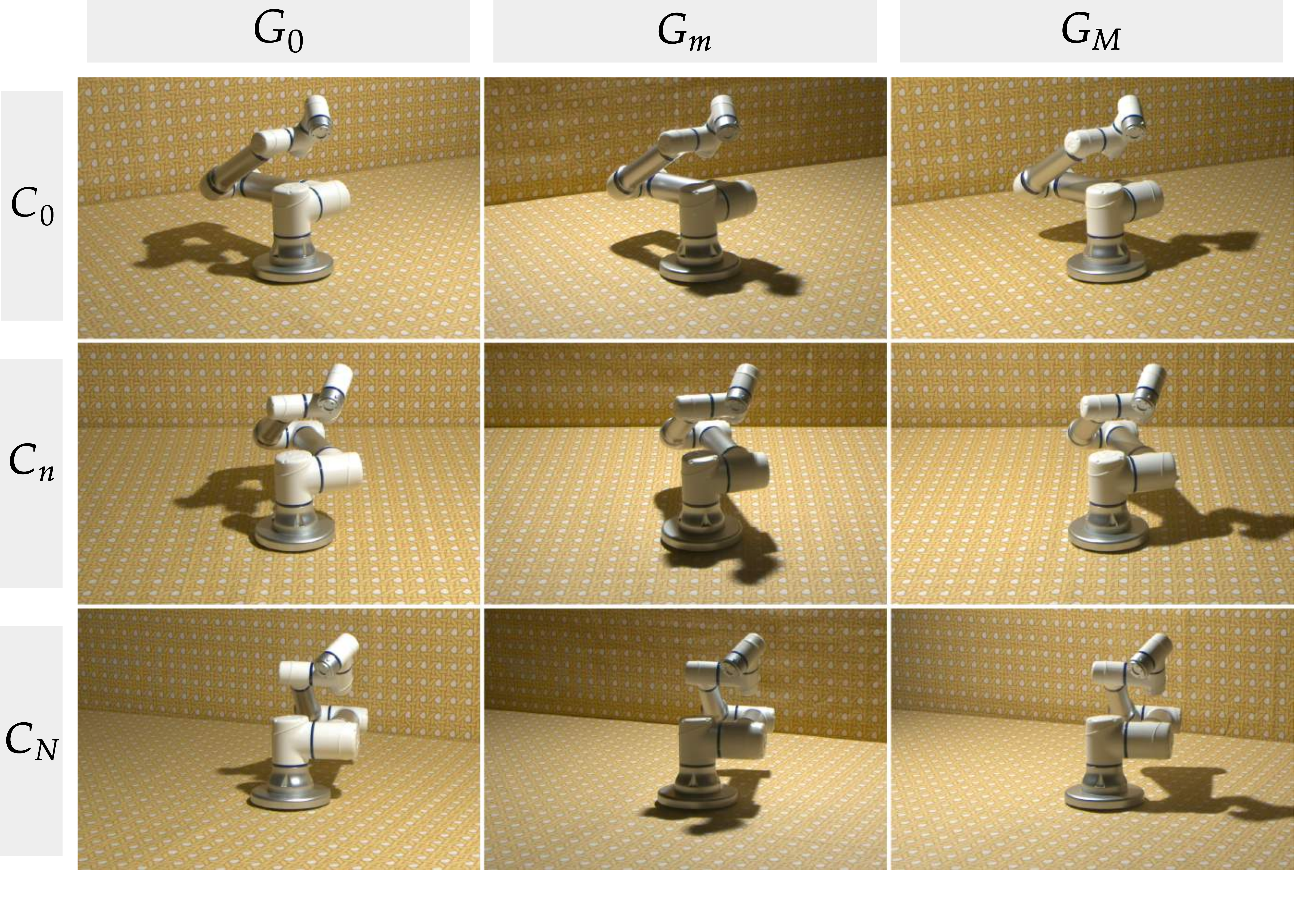}
        \caption{}
        % \caption{Grid of images in which each row depicts the same viewpoint as the light conditions change; each column, on the other hand, represents the same light condition as the viewpoint changes.}
        \label{fig:dataset_samples}
    \end{subfigure}
    \hfill
    \caption{\textbf{Overview of our dataset acquisition framework.} (a) The \lightbot{} and \camerabot{} moving light and camera respectively; the trajectories of the two robots are two non-intersecting sections of a hemisphere around the object of interest. (b) Grid of images in which each row depicts the same viewpoint as the light conditions change; each column, on the other hand, represents the same light condition as the viewpoint changes.}
    \label{fig:pipeline_dataset}
\end{figure*}
While providing both lights and cameras poses in a controlled manner with ground-truths annotations is straightforward in synthetic environments \cite{srinivasan2021nerv,zhang2021nerfactor}, it is cumbersome and complex in the real world, especially if humans are involved, as done in \cite{gross2010multi, sun2019single, wang2020single, zhang2021neural}. Our solution is to deploy extremely repeatable and safe machines such as industrial \emph{cobots} (\emph{collaborative robots}). In particular, we use a pair of robots, dubbed \lightbot{} and \camerabot{}, to position the light and camera independently. Thanks to their high repeatability (+/- \emph{0.03 mm}), we can calibrate their trajectories with respect to a common world reference frame only once before acquisition starts, and then obtain accurate 6-DoF pose knowledge of both light and camera for all scenes by repeating the same trajectories.
%Indeed, these devices have a position error of +/- \emph{0.03 mm}, which enables accurate 6-DoF pose knowledge of both light and camera in a common reference frame by means of an off-line calibration procedure. Once the two robots are calibrated, the trajectories we generate for the camera and the light can be repeated throughout all \emph{20} scenes. 
The proposed framework is sketched in \cref{fig:hardware_setup}, while some acquisitions for the Robotoy scene are shown in \cref{fig:dataset_samples}.

\noindent\textbf{Camera/Light Calibration.} In order to calibrate both robots with respect to a common reference frame we use a ChArUco board \cite{garrido2014automatic}.
We let the \camerabot{} see the board at each waypoint of its trajectory in order to calibrate extrinsics parameters for each view. Thanks to the high repeatability of the robot, it is possible to perform this procedure only once before we start to record scenes: the robot will be able to place the camera at the previously calibrated positions with negligible error. This allows us to register in a common reference frame all the camera poses for each scene without instrumenting the scene with a pattern, which alters the realism of the scene and makes the setup fragile, as the pattern may be inadvertently moved across scans. As for the \lightbot{}, we assume that the center of the point light corresponds to the geometric center of the LED. We are then able to register it with the calibration pattern by letting the end effector physically touch the central reference point of the pattern with the center of the LED at calibration time. We note we cannot use the same procedure for the \camerabot{} as the camera optical center is behind the lens. %to know the relative transformation between the position of the \lightbot{} base and the position of the pattern used to calibrate the \camerabot{}. 
That closes the calibration loop since both robots, now, will have the pattern itself as their world reference frame, thus a common coordinate system. Even with the \lightbot{}, high repeatability assures that the light pose registered at calibration time will be identical across all scenes of the dataset.

\noindent\textbf{Trajectories.} As depicted in \cref{fig:hardware_setup}, we generate two trajectories, roughly belonging to the same hemisphere, whose upper part is used for the lights while the side part for the camera. This made it possible to capture several front-facing scenes with the light moving over the object of interest. With this setup, then, for each scene we collected  50 viewpoints %$$\camera_m \in \camerasset{}$, 
each under 40 different light locations. %$\light_n \in \lightsset{}$, 
 To have waypoints uniformly distributed on the selected spherical region for each end effector, we create trajectories as sequences of centers of equal-area subregions, following the methodology proposed by \cite{kegerreis2019sea}. 
For the sake of illustration, \cref{fig:dataset_samples} shows a grid of images in which each row depicts frames with the same camera pose but different lighting conditions and vice-versa for columns.

\noindent\textbf{Background texture.} In preliminary experiments with a uniform background as scenario, we found out that NeRF struggles to correctly estimate density in such a case and training does not converge. Hence, in our dataset textured walls are shown in the background.

\noindent\textbf{Hardware.} We equipped the \lightbot{} and the \camerabot{},  respectively an Universal Robots UR5e and an Elite Robotics EC66, with a consumer headlight and an industrial camera. The headlight is the Velamp Metros IH523, equipped with 5 COB LEDs that provide a diffuse 100° beam up to 150 Lm with 6000K light temperature. We used it at half power (70 Lm) to maximize its runtime. The camera is a Basler acA1440-73gc with its optical axis mounted at 45° from the robot flange normal vector, this to simplify the computation of inverse kinematics of the \camerabot{}. Images are captured in %camera's full 
1.6MP resolution (1440px $\times$ 1080px) using a Basler lens with 8mm focal length.

\section{The \datasetname{} Dataset}
\label{sec:dataset}
In this section, we summarize the data we collected using our framework to establish the \datasetname{} dataset.
Our dataset $\sceneset{}=\{\imageset{}_{s}\}_{s=1}^{S}$ contains $S$ scenes, where each scene is a collection of RGB images $\imageset{}_{s}=\{\image_{s,n,m} \mid n=1,\dots,N, m=1,\dots,M\}$, $\image{}_{n,m} \in [0, 1]^{H \times W \times 3}$ acquired using camera poses  $\camerasset{}_{s}=\{\camera_{s,n}\}_{n=1}^{N}$ and lit with a point light source placed in $\lightsset{}_{s}=\{\light_{s,m}\}_{m=1}^{M}$. Each $\camera_n$ and $\light_m$ represent the 6D pose $\{\rotation, \translation\} \in SE(3)$ for camera and light source, respectively. $\rotation$ denotes rotation, $\rotation \in SO(3)$, and $\translation$ denotes translation, $\translation \in \mathbb{R}^3$.
The dataset features $40000$ images divided in $S=20$ scenes taken by $N=50$ frontal point of views in $M=40$ different lighting conditions, \ie{} $2000$ frames per scene.

\begin{figure}[!t]
\centering
    \includegraphics[width=0.95\linewidth, trim=0 0 0 0]{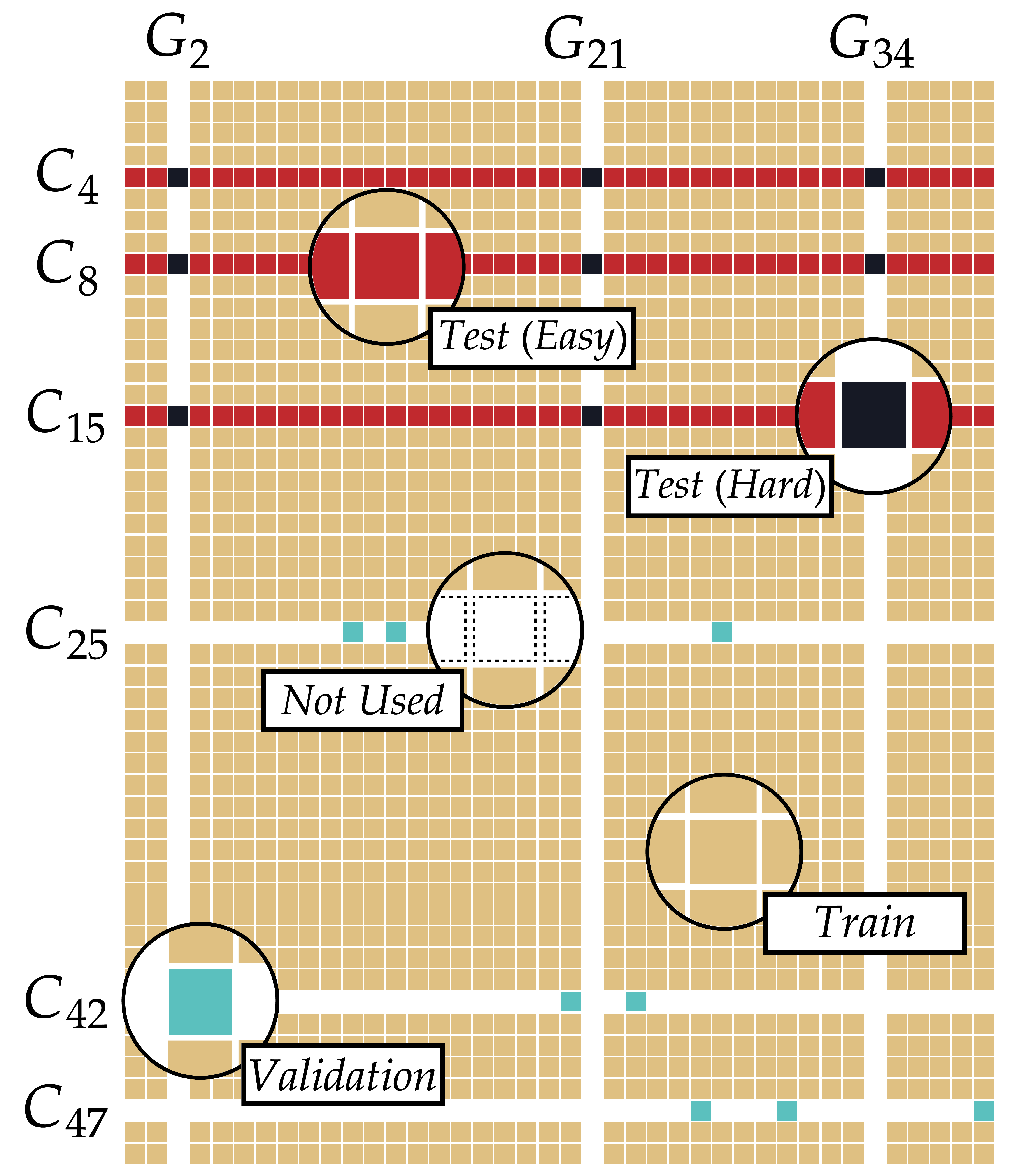}
    \caption[]{
        \textbf{Dataset splits.} Each scene $\imageset{}_{s}$ is divided in 4 different splits:\
        \begin{tikzpicture}
            \draw[fill={rgb, 255:red, 223; green, 192; blue, 130 }, line width=0pt] (0, 0) rectangle (1ex, 1ex);
            \end{tikzpicture}
        Samples used for training \
        \begin{tikzpicture}
            \draw[fill={rgb, 255:red, 91; green, 192; blue, 190 }, line width=0pt] (0, 0) rectangle (1ex, 1ex);
            \end{tikzpicture}
        Samples used for validation \
        \begin{tikzpicture}
            \draw[fill={rgb, 255:red, 193; green, 41; blue, 46 }, line width=0pt] (0, 0) rectangle (1ex, 1ex);
            \end{tikzpicture}
        Samples used for Easy Test \
        \begin{tikzpicture}
            \draw[fill={rgb, 255:red, 22; green, 25; blue, 37 }, line width=0pt] (0, 0) rectangle (1ex, 1ex);
            \end{tikzpicture}
        Samples used for Hard Test \
        \begin{tikzpicture}
            \draw[fill={rgb, 255:red, 255; green, 255; blue, 255 }, line width=0pt] (0, 0) rectangle (1ex, 1ex);
            \end{tikzpicture} Samples never used.
    } % Gives an error but works, I'm not getting it.
	\label{fig:splits}
\end{figure}

\noindent\textbf{Train, Validation and Test splits.} We split the datasets into train, val and test subsets to make it easy to compare in a fair and consistent way different approaches to NeRF relighting. 
%We use the proposed dataset in \cref{sec:benchmark} to study a novel architecture suitable for jointly novel view synthesis and relighting. As in NeRF, we train one network on each scene, $s$. 
%Each sample is, thus, an image belonging to $\imageset_{s}$ shoot from camera $\camera_{n}$ and lit by a light source $\light_{m}$, $\image_{s,n,m}$. 
In order to avoid data leakage from validation and test, we randomly create the set of held out viewpoints for validation, $\camerasset{}^{val}$, and for testing, $\camerasset{}^{test}$, as well as the set of held out light positions for testing, $\lightsset{}^{test}$. Each of them consists of $3$ indexes. To build the validation set, for each viewpoint $n_{val} \in \camerasset{}^{val}$, we randomly pick $3$ light poses among the $37$ available to form $\lightsset{}_{n_{val}}$. In this way, in the validation set we have an unseen pair (viewpoint, light) composed of a viewpoint not present in the training set and  a light used for other training viewpoints, $\imageset^{val}=\{\image_{s,n_{val},m_{val}} | n_{val} \in  \camerasset{}_{val} \wedge m_{val} \in \lightsset{}_{n_{val}} \wedge s=1,\dots,S\}$.
For test set, we build two sub-splits namely \textit{easy} and \textit{hard}. The former considers all the images $\imageset^{easy}=\{\image_{s,n,m} | n \in \camerasset{}^{test}, m=1,\dots,M, m \notin \lightsset{}^{test}\}$, while the latter is composed of images $\imageset^{hard}=\{\image_{s,n,m} | (n,m) \in \camerasset{}^{test} \times \lightsset{}^{test}\}$.
With this partition, the test set is comprised of novel viewpoints lit by light seen at training time in the easy split, and of novel viewpoints under a never seen light in the hard split. We sketch the splits in \cref{fig:splits}.

%% file: section/benchmark.tex
\section{Benchmarking the Relighting Capability of NeRF}
\label{sec:benchmark}
In the section, we explore our second research question: equipped with the \datasetname{} dataset, can we add relighting capabilities to NeRF in a simple and lightweight way? We start with a brief overview of NeRF (\cref{subsec:nerf}), we then discuss alternative ways to make NeRF able to perform  novel view synthesis and relighting simultaneously (\cref{subsec:architectures}). We compare our choices through an ablation study using the validation set described in \cref{sec:dataset}, whose results are discussed in \cref{subsec:ablation_architecture}. Finally, in \cref{subsec:results_relighting}, we present the results for the best method on both test sets of the \datasetname{} dataset.

\begin{figure*}
    \centering
    \includegraphics[width=1.\textwidth]{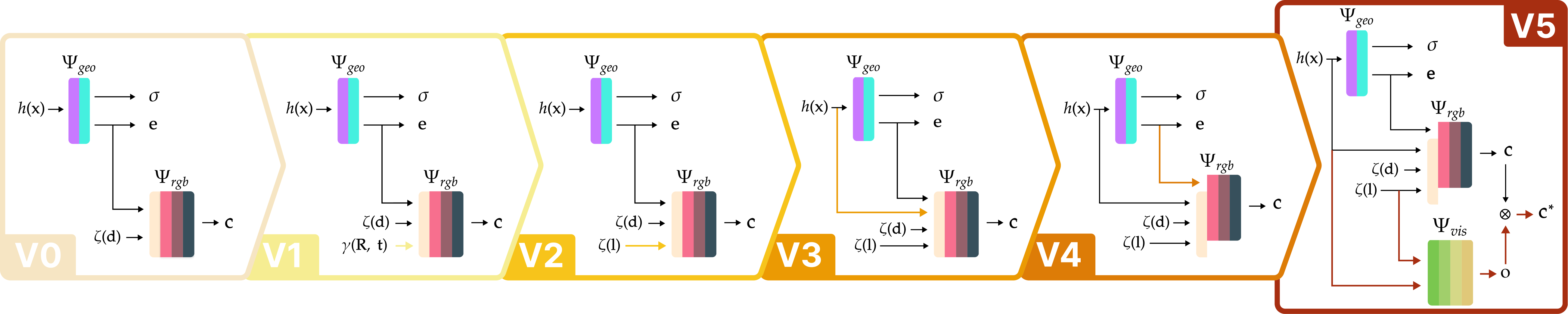}
    \caption{\textbf{Overview of architectures.} We start from the architecture of \cite{instantngp} and propose variants to inject information about the light position in the scene. Highlighted arrows in each variant underline an architectural change from the one that precedes.}
    \label{fig:architectures}
\end{figure*}

\subsection{NeRF Overview}
\label{subsec:nerf}
A Neural Radiance Field (NeRF) \cite{mildenhall2020nerf} maps a 5D input representing camera pose, \ie{} 3D coordinates $\x=(x,y,z)$ along with the 2D viewing directions $\dir=(\theta,\phi)$, into a 4D color-density output $(\mathbf{c}, \sigma)$ by means of a function $F_\Theta(\mathbf{x}, \mathbf{d}) \rightarrow (\mathbf{c}, \sigma)$ approximated by the weights of an MLP. Specifically, here we consider a vanilla NeRF architecture\cite{instantngp} which estimates color and density using two MLPs as $(\sigma, \textbf{e})$ = $\geonet(\mathbf{x})$ and $\textbf{c}=\rgbnet(\textbf{e}, \mathbf{d})$, with $\sigma$ being interpreted as the probability of a ray terminating at $(x, y, z)$ and \textbf{e} being a feature embedding. Following \cite{max1995optical}, the color $C(\mathbf{r})$ rendered from a camera ray $\mathbf{r}(t) = \mathbf{o} + t\mathbf{d}$ can be obtained solving the integral:

\begin{equation}\label{eq:rendering}
    C(\mathbf{r}) = \int_{t_n}^{t_f} T(t) \sigma(\mathbf{r}(t)) c(\mathbf{r}(t),\mathbf{d}) \textit{dt} 
\end{equation}

where $T(t) = exp \left( -\int_{t_n}^{t} \sigma (\textbf{r}(s))ds\right)$ represents the accumulated transmittance from $t_n$ to $t$ along the ray $r$, and $t$ goes from near plane $t_n$ to far plane $t_f$. The image formation procedure is performed by simply aggregating the result of the integral for all pixels, \ie{} samples along the rays, of the target image.
Although NeRF works well for synthesizing novel views, as shown by  the previous equations, it has no way of modeling the incoming light since it only takes into account position and orientation for the camera. In the next section, we will introduce variations of the classical approach that also take into account the light position.

\noindent\textbf{Positional Encoding.} As showed in NeRF \cite{mildenhall2019local} enconding $\x$ and $\dir$ into a higher dimensional space allows the neural network to overcome its spectral bias and makes it able to reproduce the high-frequency content of the input signal. We follow \cite{instantngp} and encode $\x$ using a multiresolution hash
encoding, $\hash(\x)$, whilst $\dir$ is projected onto the first 16 coefficients of the spherical harmonics, $\sh(\dir)$.

\subsection{On The Relighting Capability of NeRF}
\label{subsec:architectures}
As illustrated in \cref{fig:architectures}, we take as a starting point the architecture presented in \cite{instantngp} and delineate 5 proposals to enable NeRF to perform relighting while adding minimum complexity to the architecture. Recalling \cref{subsec:nerf}, the naive architecture, namely V0, has two sub-networks, $\Psi_{geo}$, which outputs $\sigma$ a measure of the particle density at $\textbf{x}$, and $\Psi_{rgb}$ which is responsible for the color, $\textbf{c}$. Previous works \cite{wang2021neus,wang2021neus} demonstrated that $\Psi_{geo}$ learns the scene geometry, while $\Psi_{rgb}$ models the view-dependent color appearance. Hence, our key idea is to modify the input to the $\rgbnet$ injecting light position and let the model learn the interaction between the geometry and light. We train one network on each scene minimizing the error between the ground-truth images and the rendered ones. As input data we consider a set of multi-view images of an object, $\imageset{}_{s}$, illuminated under known lighting conditions, $\lightsset{}_{s}$, and the camera poses of these images, $\camerasset{}_{s}$. We will consider the subscript $s$ implicit in what follows.

\noindent\textbf{V1.} A straightforward way to enable relighting with NeRF is to condition $\Psi_{rgb}$ with a latent representation that stores the world position and orientation of the light, $\light{}_{m}$ for a given observation $\image_{n,m}$. To do so, we flatten the $\rotation$ and $\translation$ parts of $\light{}_{m}$ and encode this 12-dimensional vector into a 156-dimensional latent code using the positional encoding with Fourier Features proposed in \cite{mildenhall2020nerf}, $\gamma(\rotation,\translation)$. Our procedure resembles the approach adopted by NeRF in the Wild \cite{martin2021nerf}, but the condition embedding is used to adapt NeRF to variable lighting conditions instead of transient parts of the images.
% Previous image-based techniques \cite{xu2018deep,sun2019single} often send lighting information as additional inputs to the network, and compute challenging view- or light-dependent shading effects through the network processing. 
% change how we determine the input light information that for each queried 3D point is passed (??) to the color MLP, alongside the view direction. Both geometry and color MLP architectures stay the same. \newline

\noindent\textbf{V2.} As the next iteration, we change the parameterization for $\light{}_{m}$. Instead of directly encoding $\rotation$ and $\translation$ in the same way for each sample $\x$ used to query $\rgbnet{}$, we construct a 3D unit vector $\ldir=\frac{\translation-\x} {||\translation-\x||}$ which let the network be aware of the relative position between the light source and the query location. Since, differently from the previous approach, $\ldir$ changes for each query point, we conjecture this may help NeRF in modulating the output color $\mathbf{\cl}$ based on the incoming light information. As done for $\dir$, $\ldir$ is encoded using the spherical harmonics, $\sh(\ldir)$.

% In fact, the light pose (??) is defined per image and thus is fixed for every 3D point along every ray casted from each pixel on the image towards the scene, while the opposite is true for the $\omega_{i}$ vector, which changes for every point. \newline
% In this way both the viewing direction and the light direction are expressed as 3D Cartesian unit vectors, and are both mapped to a higher dimensional space before being fed to the color MLP. 

\noindent\textbf{V3.} The previous version resembles somehow the BRDF where $\ldir$ acts as the incoming light direction at $\x$, and $\mathbf{d}$ is the unit vector pointing from $\x$ toward the camera. Following this analogy, we may consider $\geoemb{}$ to represent an $N$-dimensional embedding of the normal vector at $\x$. Since all these information are relative to $\x$, we explore whether having $\hash(\x)$ as input can be beneficial for $\rgbnet$. Without such skip connection the knowledge about the location of the query point may vanishes as the depth of the network increases. 

\noindent\textbf{V4.} The previous version may be negatively influenced by the diverse scales of its inputs, which are on one side spherical and hash-grid encoded vectors like $\sh(\mathbf{d})$, $\sh(\ldir)$, and $\hash(\x)$, while on the other there is the ReLU output $\geoemb$. To test this hypothesis, we propose a variant where we concatenate $\geoemb$ with the output of the first layer of $\rgbnet$ and use this vector as input for its second layer, rather than feeding $\rgbnet$ directly with $\geoemb$. %In this way, we avoid to use as input for $\rgbnet$ a combination of vectors with mixed magnitude, \eg{} spherical and hash-grid encoded vector $\sh(\mathbf{d})$, $\sh(\ldir), \hash(\x)$, and ReLU activated outputs of linear layer, \eg{} $\geoemb$.
%As stated before, the original NeRF models the scene as a continuous 3D field of particles that absorb and \textit{emit} light. In fact, only the amount of outgoing light from a location is modeled: the fact that this outgoing light is the result of interactions between incoming light and the material properties of an underlying surface is ignored. This means that the learned representation does not separate the effect of incident light from material properties of surfaces, with the material and lighting effects baked in, which makes them not editable. \newline

\noindent\textbf{V5.} The original NeRF learns a continuous 3D field of particles that absorb and \textit{emit} light. As a result only the amount of outgoing light from a location is modeled, without taking into account the fact that the outgoing light is the result of interactions between incoming light and the material properties of an underlying surface. This issue can be alleviated by replacing \cref{eq:rendering} with a physically-based volume rendering for non-emissive and non-absorptive volumes, which replaces the emitted color of each 3D point along the ray, $c(\mathbf{r}(t),\mathbf{d})$, with $L_{r}(\x,\omega_{o})$ that represents the scattered light at $\x$ along $\omega_{o}$. As done in \cite{bi2020neural}, assuming a single point light source of fixed intensity, $L_{r}$ can be approximated as:
\begin{equation}
    L_{r}(\x,\omega_{o}) =\int_{S} f_{p}(\x,\omega_{o},\omega_{i})L_{i}(\x,\omega_{i})d\omega_{i}
\end{equation}
 where $S$ is a unit sphere, $f_{p}$ represents a differentiable reflectance model (like the BRDF), and $L_{i}$ represents the incident radiance at $x$ from direction $\omega_{i}$. Please note that in this equation $\omega_{o}$ correspond to $\mathbf{d}$ and $\omega_{i}$ is analogous to $\ldir$.
While we can assume that $f_{p}$ is approximated by $\rgbnet$, computing $L_{i}$ requires marching a large number of light rays for all shading points on all camera rays to determine the transmittance between the light and each 3D point. We avoid this cumbersome procedure by introducing a neural approximation of $L_{i}$ in the form of an MLP aimed at predicting a scalar value $o=\shanet( \hash(\x),\sh(\ldir))$, which allows us to efficiently query the point-to-light visibility. The final pixel color is then obtained with $\mathbf{\cl}^{\star} = o \cdot \mathbf{\cl}$.

\subsection{Ablation study}
\label{subsec:ablation_architecture}
\noindent\textbf{Dataset.} To assess the relative merits of the incremental modifications to NeRF proposed in the previous section and identify the best architecture, we consider a subset of scenes presenting a variety of challenges: Cube, that has thin wireframe structures and shadows; Savannah, that pictures a jagged leaves tree with a complex shadow; Reflective, that is full of specular reflections, and FlipFlop, where light passes through semi-transparent plastic strings.

\noindent\textbf{Implementation Details.} 
We used the torch-ngp pytorch implementation\footnote{\url{https://github.com/ashawkey/torch-ngp}} of Nvidia Instant-NGP \cite{instantngp} for fast training of Neural Radiance Fields models. During training, we randomly sample 4096 pixel rays as a batch to train our networks. We use Adam optimizer \cite{adam} with an initial learning rate of 0.01 (other Adam hyperparameters are left at default values of $\beta_{1} = 0.9$, $\beta_{2} = 0.999$).  For volumetric integration, we use 128 samples in coarse volume and 128 additional adaptive samples in fine volume to compute the final radiance. The optimization for a single scene typically takes around 100–200k iterations to converge on a single NVIDIA RTX 2080Ti GPU (about 5 hours). We use early-stopping if the model doesn't improve for 10 consecutive epochs. We supervise the regressed RGB values with the ground truth values from the captured images using the L2 loss.
For $\geonet{}$ we use 2 fully-connected ReLU layers with 64 channels, while $\rgbnet{}$ and 
$\shanet{}$ uses 4 fully-connected ReLU layers with 64 channels. 

% Instead of directly passing 3D coordinates x, camera direction vectors $\omega_{o}$ and light direction vectors $\omega_{i}$ to the MLPs, we map these inputs using positional encoding. While the 3D position is mapped via Multiresolution Hash Encoding from Instant-NGP paper \cite{instantngp}, camera and light directions are mapped via Spherical Harmonics. 

\noindent\textbf{Results.} For a quantitative evaluation of our methods we consider two widely adopted metrics of image quality assessment: Structural Similarity (SSIM) \cite{ssim} and Peak Signal-to-Noise Ratio (PSNR) \cite{mildenhall2019local}. We report the results of the ablation study in \cref{tab:architecture_ablation}. Results show how all the proposed modifications to NeRF improve performance and contribute to the overall good results of the most effective version, which is V5. These results provide experimental support to the intuitions presented above about the higher effectiveness of feeding to the network the relative light position with respect to the sample $\x$ instead of the absolute light pose (row 1 versus row 2), as well as the importance of providing $\x$ as input to $\rgbnet{}$ (row 2 versus row 3). Separating inputs to $\rgbnet{}$ is also beneficial and, together with $\shanet{}$ contributes to the good performance of V5.

\input{results/ablation_architecture.tex}

Some qualitative examples comparing and contrasting V1 against V5 are reported in \cref{fig:quali_ablation}.
Zooming in on cast shadows we can appreciate how careful handling of the light pose as well as the proposed modifications to the vanilla NeRF architecture enable rendering of sharper and more coherent shadows than the straightforward extension that conditions NeRF also on light pose. %Thus, as can be seen, a more structured architecture, such as V5, is able to represent shadows much more sharply because of the decoupling between the albedo component and the visibility part of each point with respect to the light source. 

\input{tables/quali_ablation.tex}

\subsection{Results on benchmark}
\label{subsec:results_relighting}
Finally, we report the performance for the best architecture on the overall set of held out test scenes in \cref{tab:relight}. We compare our method against the pytorch implementation of Nvidia Instant-NGP \cite{instantngp}. Overall, the proposed architecture achieves satisfying results on this challenging task and sets a non trivial baseline for the online benchmark, clearly outperforming the naive baseline represented by the standard NeRF.
We show some qualitative results for the easy test split in \cref{fig:quali_easy}, and in \cref{fig:quali_hard} for the hard one. These qualitative results confirm that the model is capable of convincingly rendering and lighting the scene from unseen viewpoints and light positions while using a simple and lightweight architecture. Interestingly, the output of $\shanet{}$, reported in the rightmost columns of the figures, well approximate point-to-light visibility, especially in the easy split, even if it has not received direct supervision to emulate it, and it even reproduces some indirect illumination effects, confirming the feasibility of a neural approximation of the incident radiance $L_i(\x, \omega_i)$. Comparing with the ground-truth frame, while rendering of light on the object is indeed quite realistic, cast shadows are sharp and coherent with the scene geometry on the easy set, while more artifacts are present in the hard split.

%Although One can see a slight decrease in metrics on the \emph{Hard} test set, compared to the \emph{Easy} test set, as expected. The difference is not that substantial because probably the background pattern dominates over the image than the object itself. 
\input{results/relight_cr}
% commenti ai qualitativi
% riproduce indirect illumination, approximating integral with MLP is effective.

% non sharp boundaries of shadows
% texture copy artifact
% 
\input{tables/quali_easy.tex}

\input{tables/quali_hard.tex}

%% file: results/ablation_architecture.tex
\begin{table}
    \resizebox{\linewidth}{!}{
      \centering
       \setlength{\tabcolsep}{1.5mm}
      {\small
       \begin{tabular}{l|ccccc|cccccccc}
       \toprule
       Method & $\gamma(\rotation,\translation)$ & $\ldir$ & \textit{skip} & \textit{inputs} & $\Psi_{vis}$  & \multicolumn{2}{c}{Cube} & \multicolumn{2}{c}{Savannah} & \multicolumn{2}{c}{Reflective} & \multicolumn{2}{c}{FlipFlop} \\
       \midrule
       & & & & & & PSNR & SSIM & PSNR & SSIM & PSNR & SSIM & PSNR & SSIM \\  
       \midrule
       V1 & \cmark & \xmark & \xmark & \xmark & \xmark & 24.37 & 0.52 & 22.53 & 0.44 & 23.57 & 0.51 & 24.12 & 0.51 \\
       V2 & \xmark & \cmark & \xmark & \xmark & \xmark & 24.73 & 0.54 & 23.70 & 0.52 & 23.68 & 0.52 & 24.42 & 0.56 \\
       V3 & \xmark & \cmark & \cmark & \xmark & \xmark & 25.38 & 0.56 & 24.39 & 0.55 & 24.65 & 0.58 & 25.06 & 0.57 \\
       V4 & \xmark & \cmark & \cmark & \cmark & \xmark & 25.41 & 0.57 & 24.79 & 0.58 & 24.24 & 0.56 & 25.27 & 0.58 \\
       V5 & \xmark & \cmark & \cmark & \cmark & \cmark & \textbf{26.11} & \textbf{0.61} & \textbf{25.23} & \textbf{0.61} & \textbf{25.00} & \textbf{0.59} & \textbf{25.46} & \textbf{0.60} \\
       \bottomrule
      \end{tabular}
      }
   }
  %\vspace{3mm}
  \caption{\label{tab:architecture_ablation} Quantitative relighting and view synthesis result of the proposed modifications to NeRF. For each scene, we report on the left the PSNR and on the right the SSIM. Legend: $\gamma(\rotation,\translation)$ indicates the use of absolute light position, $\ldir$ of relative position wrt $\x$, \textit{skip} indicates that $\x$ is provided as input to $\rgbnet{}$, \textit{inputs} that the embedding \textbf{e} is inserted in $\rgbnet{}$ at the second layer, while $\Psi_{vis}$ that the MLP approximating $L_i(x, \omega_i)$ is used. }
\end{table}

%% file: tables/quali_ablation.tex
\begin{figure}
    \centering
    % \begin{subfigure}[b]{0.23\textwidth}
    %     \centering
    %     \includegraphics[width=\textwidth]{images/quali/ablation_quali/quali_zooms_v1.pdf}
    %     \caption{}
    %     \label{fig:quali_ablation_v1}
    % \end{subfigure}
    % % \hfill
    % \begin{subfigure}[b]{0.23\textwidth}
    %     \centering
    %     \includegraphics[width=\textwidth]{images/quali/ablation_quali/quali_zooms_v5.pdf}
    %     \caption{}
    %     \label{fig:quali_ablation_v5}
    % \end{subfigure}
    % % \hfill
    \includegraphics[width=0.48\textwidth]{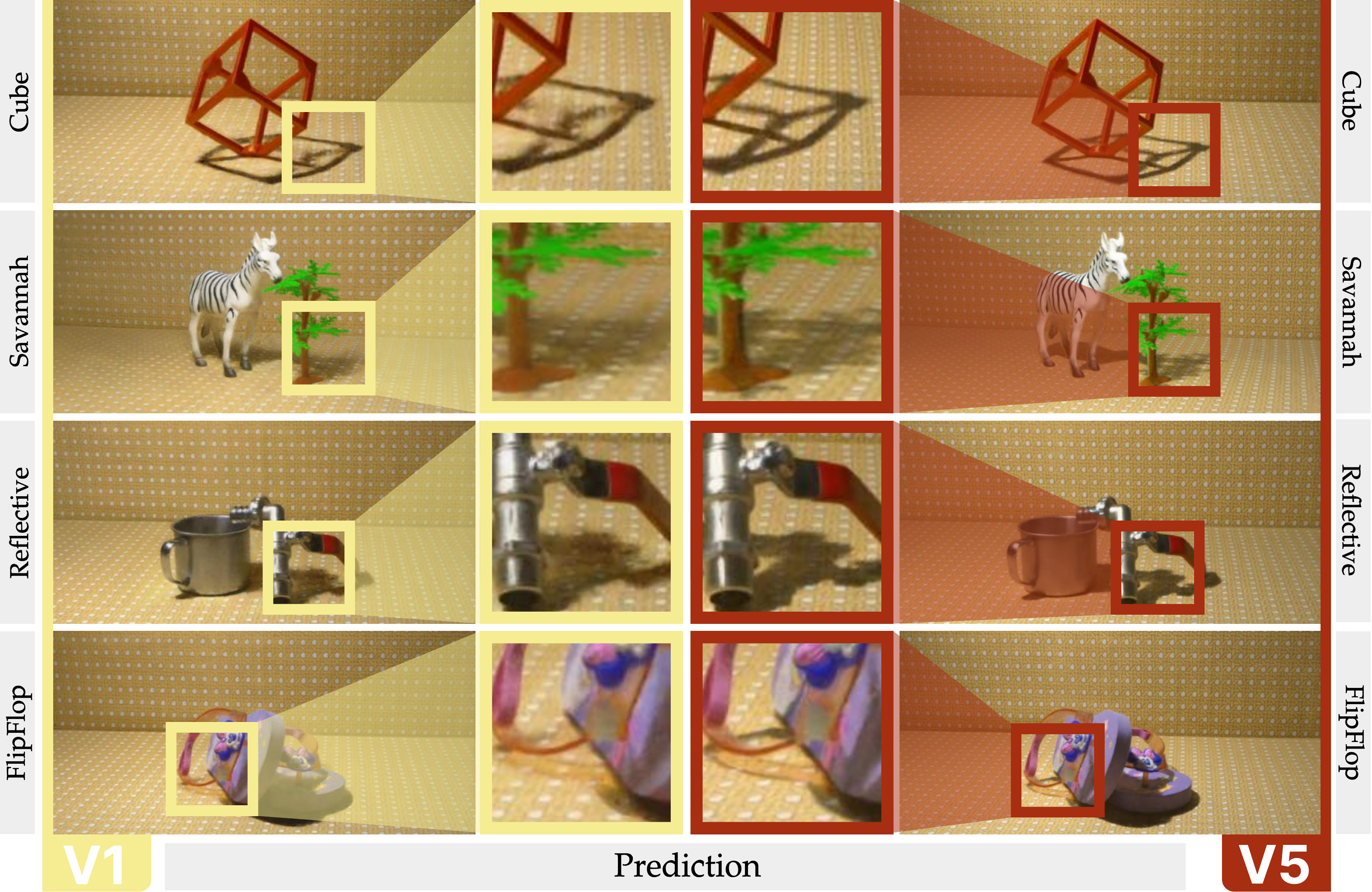}
    \caption{\textbf{Qualitative results for ablation.} The same image as rendered by V1 and V5 networks. The side-by-side comparison clearly shows how V5 reproduces much sharper details in shadows (best seen in Reflective) and is able to better handle complex reflections (best seen in FlipFlop).}
    \label{fig:quali_ablation}
\end{figure}

%% file: results/relight_cr.tex
\begin{table}
	\centering
	\resizebox{1.0\linewidth}{!}{
		\setlength{\tabcolsep}{1.5mm}
		{\small
			\begin{tabular}{l|cc|cc|cc|cc}
				\toprule
				\multicolumn{1}{c}{} &
				\multicolumn{4}{c}{\textit{Ours}} &
				\multicolumn{4}{c}{\textit{Instant-NGP}} \\
				\toprule
				\multicolumn{1}{c}{Name} &
				\multicolumn{2}{c}{\textit{Easy}} &
				\multicolumn{2}{c}{\textit{Hard}} &
				\multicolumn{2}{c}{\textit{Easy}} &
				\multicolumn{2}{c}{\textit{Hard}} \\
				\midrule
				& PSNR & SSIM & PSNR & SSIM & PSNR & SSIM & PSNR & SSIM \\       
				\hline                     
				Apple 		& 26.44 & 0.62 & 26.25 & 0.62 & 20.89 & 0.45 & 20.95 & 0.45 \\
				Cheetah 	& 25.66 & 0.61 & 24.64 & 0.60 & 19.37 & 0.44 & 19.60 & 0.44 \\
				Cube 		& 24.90 & 0.54 & 23.98 & 0.53 & 20.14 & 0.42 & 20.31 & 0.42 \\
				Dinosaurs 	& 25.75 & 0.65 & 24.98 & 0.64 & 19.58 & 0.42 & 19.66 & 0.41 \\				  
				FlipFlop 	& 25.85 & 0.61 & 25.42 & 0.61 & 20.38 & 0.45 & 20.36 & 0.45 \\
				Fruits 		& 25.93 & 0.62 & 25.72 & 0.62 & 20.16 & 0.45 & 20.21 & 0.44	\\
				Garden 		& 25.74 & 0.66 & 25.08 & 0.66 & 19.76 & 0.45 & 19.70 & 0.45	\\
				Helicopters & 25.12 & 0.61 & 24.73 & 0.61 & 19.34 &	0.37 & 19.37 & 0.37	\\
				Kittens 	& 25.90 & 0.64 & 24.96 & 0.63 & 18.52 &	0.37 & 18.65 & 0.37	\\			
				Lego 		& 26.07 & 0.61 & 25.77 & 0.61 & 20.75 &	0.46 & 20.76 & 0.46	\\			
				Lunch 		& 25.84 & 0.60 & 24.71 & 0.59 & 19.32 &	0.46 & 19.38 & 0.45	\\			
				Plant 		& 26.55 & 0.67 & 25.93 & 0.67 & 20.62 &	0.44 & 20.66 & 0.44 \\				
				Reflective 	& 25.79 & 0.61 & 25.28 & 0.61 & 20.09 &	0.43 & 20.11 & 0.42	\\			
				Robotoy 	& 26.24 & 0.65 & 25.55 & 0.65 & 20.77 &	0.50 & 20.78 & 0.50	\\
				Savannah 	& 25.15 & 0.62 & 24.31 & 0.61 & 19.08 &	0.40 & 19.18 & 0.40	\\			 
				Shark 		& 25.59 & 0.57 & 25.32 & 0.56 & 20.54 &	0.42 & 20.53 & 0.41	\\			
				Stegosaurus & 25.87 & 0.63 & 25.65 & 0.63 & 20.84 &	0.43 & 20.91 & 0.42	\\			
				Tapes 		& 25.84 & 0.58 & 25.41 & 0.57 & 19.34 &	0.41 & 19.55 & 0.41	\\			        
				Trucks 		& 25.80 & 0.67 & 25.16 & 0.66 & 19.81 &	0.44 & 19.87 & 0.44	\\			    
				Wooden toys & 25.69 & 0.61 & 25.24 & 0.60 & 20.19 &	0.48 & 20.22 & 0.48	\\			
				\midrule
				Average 	& \textbf{25.79} & \textbf{0.62}  & \textbf{25.20} & \textbf{0.61}  & 19.97 & 0.43 & 20.04 & 0.43 \\
				\bottomrule
			\end{tabular}
		}
	}
	%\vspace{3mm}
	\caption{\label{tab:relight} Quantitative results across all 20 scenes. We report both Peak Signal-to-Noise Ratio (PSNR) and Structural Similarity (SSIM) by comparing the novel view synthetized image with its corresponding groundtruth.}
\end{table}

%% file: tables/quali_easy.tex
% \begin{table}[h]
% \resizebox{\linewidth}{!}{
% \begin{tabular}{ccc}
% \includegraphics[width=0.30\linewidth,trim={150 275 200 0},clip]{images/quali/easy_quali/cube/0037_gt.png} 
% &\includegraphics[width=0.30\linewidth,trim={150 275 200 0},clip]{images/quali/easy_quali/cube/0037_image.png}
% &\includegraphics[width=0.30\linewidth,trim={150 275 200 0},clip]{images/quali/easy_quali/cube/0037_shadows.png}\\
% \newline
% \includegraphics[width=0.30\linewidth,trim={250 250 100 25},clip]{images/quali/easy_quali/kittens/0065_gt.png} 
% &\includegraphics[width=0.30\linewidth,trim={250 250 100 25},clip]{images/quali/easy_quali/kittens/0065_image.png}
% &\includegraphics[width=0.30\linewidth,trim={250 250 100 25},clip]{images/quali/easy_quali/kittens/0065_shadows.png}\\
% \newline
% \includegraphics[width=0.30\linewidth,trim={200 275 150 0},clip]{images/quali/easy_quali/reflective/0071_gt.png}
% &\includegraphics[width=0.30\linewidth,trim={200 275 150 0},clip]{images/quali/easy_quali/reflective/0071_image.png}
% &\includegraphics[width=0.30\linewidth,trim={200 275 150 0},clip]{images/quali/easy_quali/reflective/0071_shadows.png}\\
% \end{tabular}
% \label{fig:quali_easy}
% }
% \caption{From left to right we show the ground truth, predicted image and shadow map for three scenes}
% \end{table}

\begin{figure}[b]
    \centering
    \includegraphics[width=0.5\textwidth]{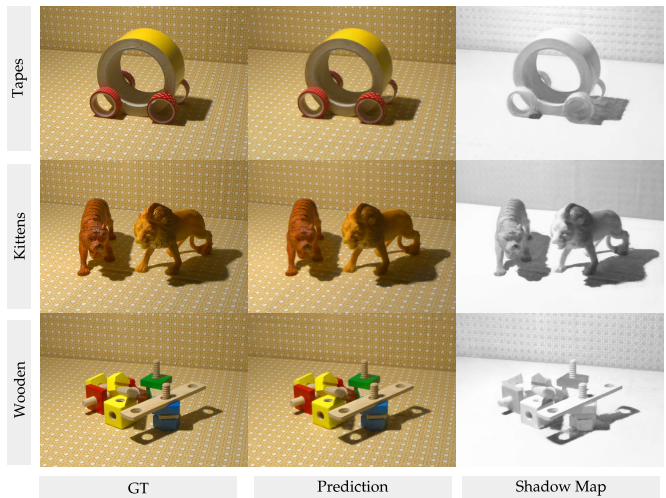}
    \caption{We show one image from the easy test split $\imageset^{easy}$ on the left. In the middle we can see the same image as rendered by our V5 model, while on the right the intermediate visibility output of $\shanet{}$ is shown, which is then multiplied by the predicted BRDF to determine the outgoing radiance at each point.}
    \label{fig:quali_easy}
\end{figure}

%% file: tables/quali_hard.tex
% \begin{table}[h]
% \resizebox{\linewidth}{!}{
% \begin{tabular}{ccc}
% \includegraphics[width=0.30\linewidth,trim={300 400 300 50},clip]{images/quali/hard_quali/apple/0004_gt.png} 
% &\includegraphics[width=0.30\linewidth,trim={300 400 300 50},clip]{images/quali/hard_quali/apple/0004_image.png}
% &\includegraphics[width=0.30\linewidth,trim={300 400 300 50},clip]{images/quali/hard_quali/apple/0004_shadows.png} \\
% \newline
% \includegraphics[width=0.30\linewidth,trim={300 400 300 50},clip]{images/quali/hard_quali/plant/0004_gt.png} 
% &\includegraphics[width=0.30\linewidth,trim={300 400 300 50},clip]{images/quali/hard_quali/plant/0004_image.png}
% &\includegraphics[width=0.30\linewidth,trim={300 400 300 50},clip]{images/quali/hard_quali/plant/0004_shadows.png} \\
% \newline
% \includegraphics[width=0.30\linewidth,trim={250 300 350 150},clip]{images/quali/hard_quali/stegosaurus/0007_gt.png}
% &\includegraphics[width=0.30\linewidth,trim={250 300 350 150},clip]{images/quali/hard_quali/stegosaurus/0007_image.png}
% &\includegraphics[width=0.30\linewidth,trim={250 300 350 150},clip]{images/quali/hard_quali/stegosaurus/0007_shadows.png} \\
% \end{tabular}
% \label{fig:quali_hard}
% }
% \caption{}
% \end{table}

\begin{figure}[b]
    \centering
    \includegraphics[width=0.5\textwidth]{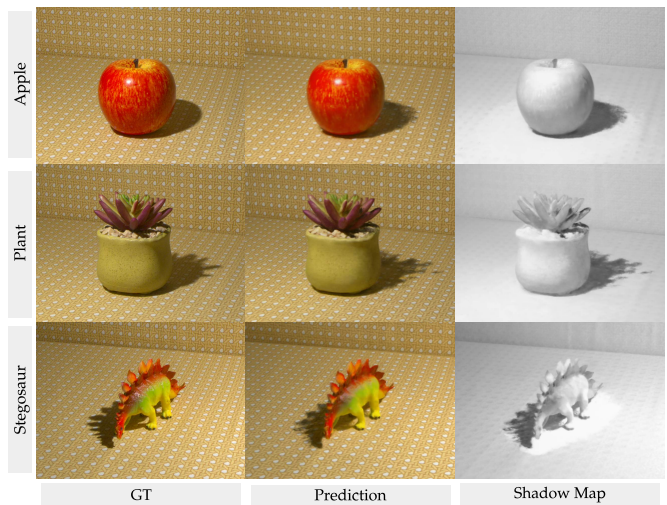}
    \caption{On the left the ground-truth image drawn from the hard test split $\imageset^{hard}$. In the middle the corresponding render produced by our V5 model. On the right the intermediate visibility output of our model.}
    \label{fig:quali_hard}
\end{figure}

%% file: section/conclusion.tex
\section{Conclusion and Limitations}
\label{sec:conclusion}
We have introduced the \datasetname{} dataset, the first dual robot dataset framing real world objects under challenging one-light-at-time (OLAT) conditions and annotated with accurate camera and light poses. The main limitations of our dataset concern the absence of 360 degrees scans and the use of a challenging but unrealistic OLAT setup. % We will release it upon publication, together with an online evaluation server that will wrap the held out test images of each scene and will enable a fair comparison of future methods. 
By leveraging the training and validation splits of the dataset, we were able to perform an ablation study on lightweight modifications to a NeRF architecture that extend it to successfully perform novel view synthesis under unseen lighting. The best model emerged from the study has been tested on the held out test set to establish a non-trivial baseline for the benchmark. We hope the availability of a new dataset and baseline for the problem of NeRF relighting will attract new research around this challenging inverse rendering problem.